\documentclass[conference]{IEEEtran}
\usepackage{cite}
\usepackage{array}
\usepackage{cite}
\usepackage{longtable}
\usepackage{amsmath}
\usepackage{epsfig}
\usepackage{algorithmic}
\usepackage{algorithm}

\begin{document}
%
\title{A cognitive diversity framework for radar target classification}

\author{\authorblockN{Amit K. Mishra}
\authorblockA{Department of ECE\\
Indian Institute of Technology Guwahati\\
Guwahati, India\\
Email: akmishra@ieee.org}
\and
\authorblockN{Chris J. Baker}
\authorblockA{College of Engineering and\\Computer Science\\
Australian National University\\
Email: chris.baker@anu.edu.au}}

\maketitle

\begin{abstract}
Classification of targets by radar has proved to be notoriously difficult with the best systems still yet to attain sufficiently high levels of performance and reliability. 
In the current contribution we explore a new design of radar based target recognition, where angular diversity is used in a cognitive manner to attain better performance. 
Performance is bench-marked against conventional classification schemes.
The proposed scheme can easily be extended to cognitive target recognition based on multiple diversity strategies. 
\end{abstract}

\section{Introduction}
Even after almost two decades of research, radar based Automatic Target Recognition (ATR) is still a challenging engineering problem for which a robust solution remains elusive. Diversity in space, time, frequency and waveform offers new design freedoms yet to be fully explored. 
Indeed, multi-perspective radar based ATR, exploiting spatial diversity has been shown to offer a significant improvement in performance \cite{vespe_multi}. 
This improvement is based on the fact that radar returns from a target are generally highly aspect dependent and new information can be gleaned that provides a more unique overall target descriptor.

More recently still, the concept of cognition has been advocated as offering new and improved performance for radar in a variety of ways \cite{haykin_06}.  Synthetic cognition emerges from two major considerations. Firstly, the sensor parameters to be transmitted should be variable so that they can be altered to acquire the most useful information. Secondly, the setting of those parameters and collection of sensor data should be performed based on prior information regarding the object or scene under observation as well as the sensor data collected in previous time intervals. Of course this is all done to achieve a desired goal.

However, it may be marked here that the definition of cognitive behaviour and when can a system be called cognitive, are still areas of hot debate \cite{fred_10}. 
Hence, what we are really doing is to take different clues from cognitive and bio-inspired system literature and applying them in the field of radar system engineering in an endeavour to enhance the performance. 
In a strict sense, such system can at best be called smart, bio-inspired or adaptive. 

In this paper we describe a strategy for cognitive diversity ATR using radar based sensors. The primary diversity parameters are waveform and angle. The cognitive processing approach is developed from an analysis of the neurological observations of echo-locating mammals \cite{bat}. 
Detailed structure of the proposed processing architecture is discussed in section II. 

Rest of the paper is organised as follows. 
Section II discusses the database used in the current study. 
Section III explains the cognitive ATR processor framework proposed in this work. 
It also discusses the two fine-tuned versions of the basic cognitive ATR frameworks used in validating the overall scheme. 
The next section discusses the results. The last section concludes the paper with some discussions on the aspects of the strategy to be explored in future.

\section{Database used}
\begin{figure}
\centering
	\includegraphics[width=110mm]{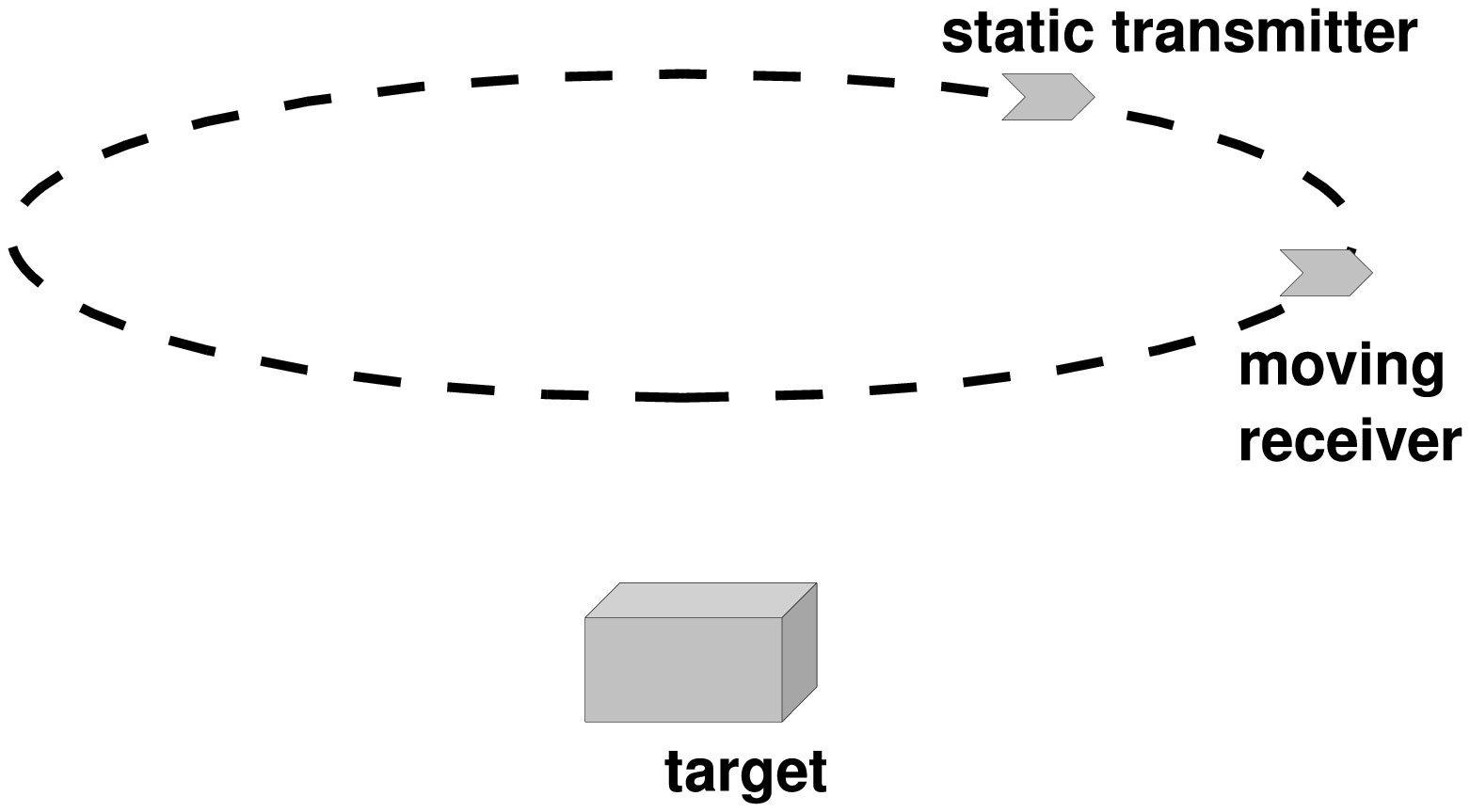}
	\caption{Transmitter and receiver configuration in the simulation setup} \label{txrx}
\end{figure}

\begin{figure*}[htbp]
\begin{center}
$\begin{array}{c@{\hspace{0.1in}}c}
\multicolumn{1}{l}{\mbox{\bf }} &
        \multicolumn{1}{l}{\mbox{\bf }} \\ [-0.13cm]
\epsfxsize=2.5in
\epsffile{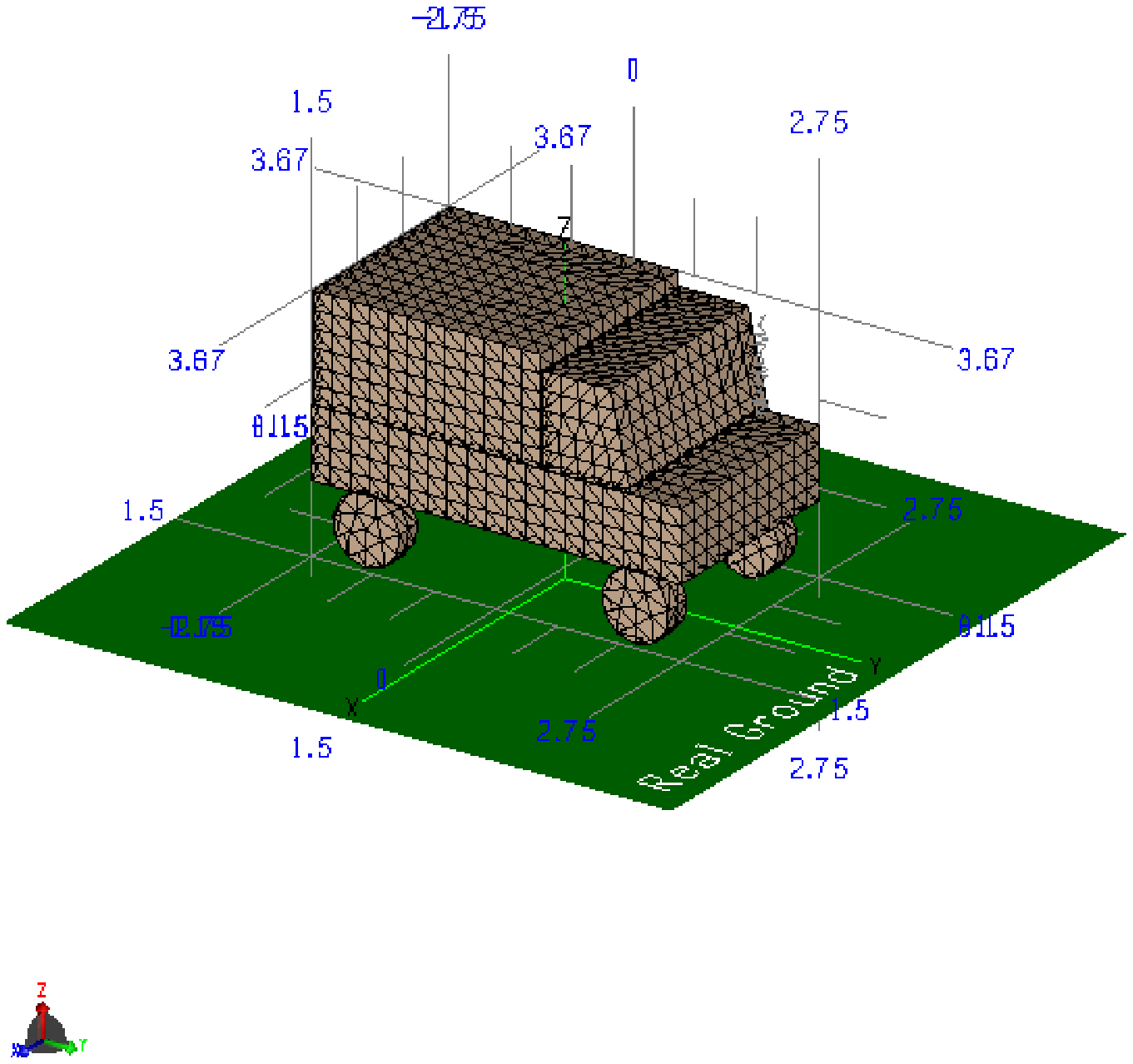} &
        \epsfxsize=2.5in
        \epsffile{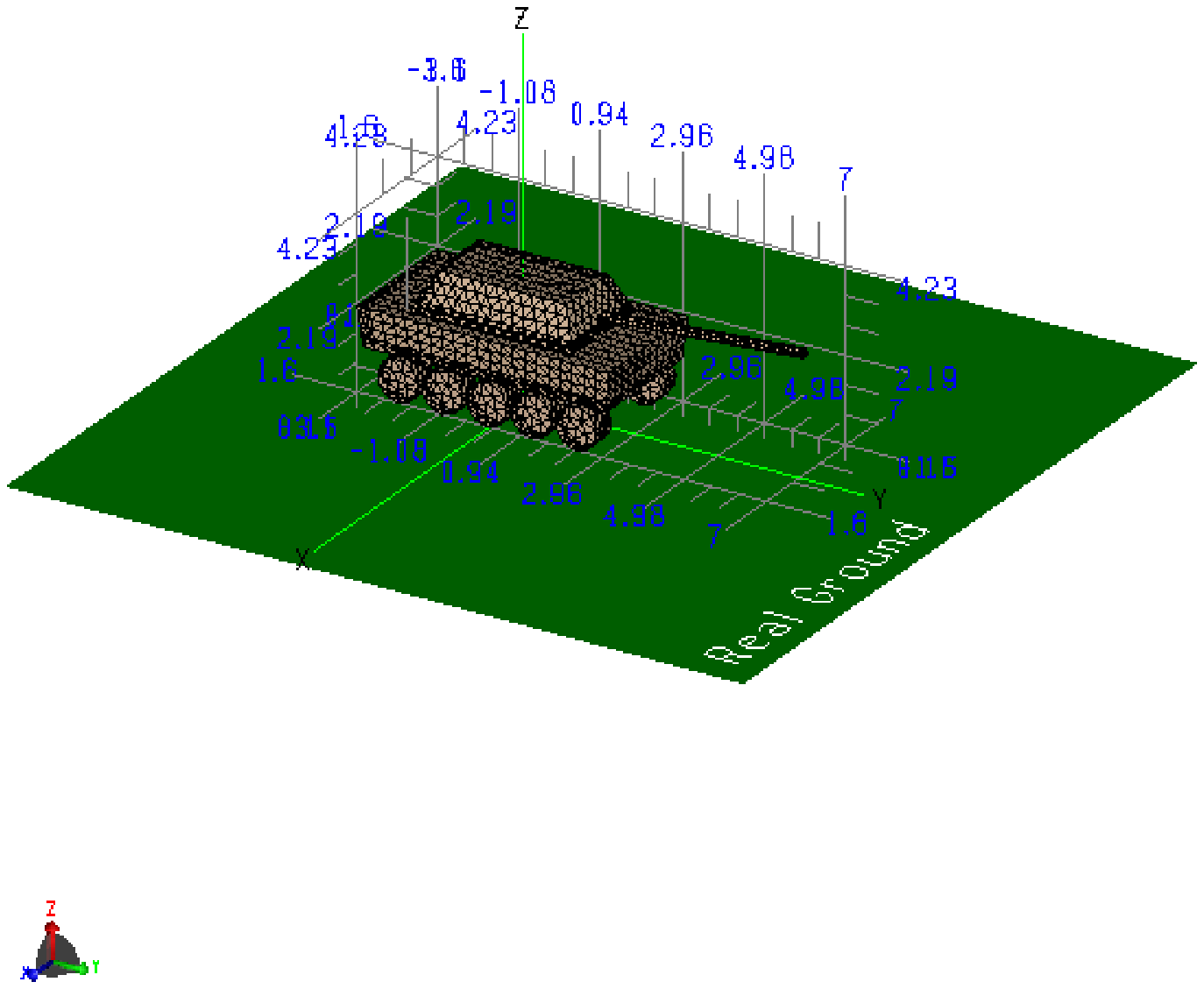} \\ [0.1cm]
\mbox{\bf Armored personal carrier (APC)} & \mbox{\bf Main battle tank (MBT)}
\\
\multicolumn{1}{l}{\mbox{\bf }} &
        \multicolumn{1}{l}{\mbox{\bf }} \\ [-0.13cm]
\epsfxsize=2.5in
\epsffile{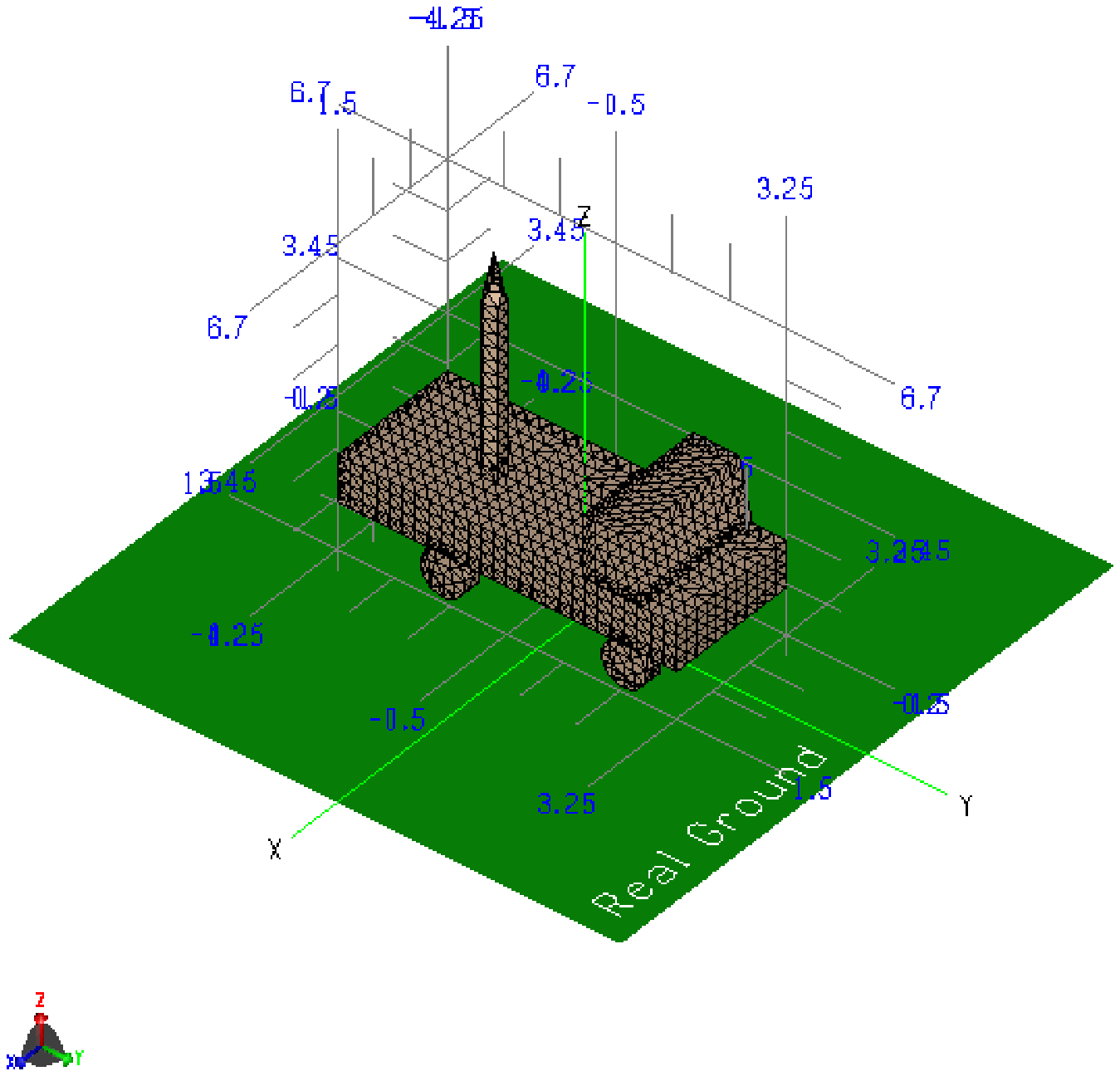} &
        \epsfxsize=2.5in
        \epsffile{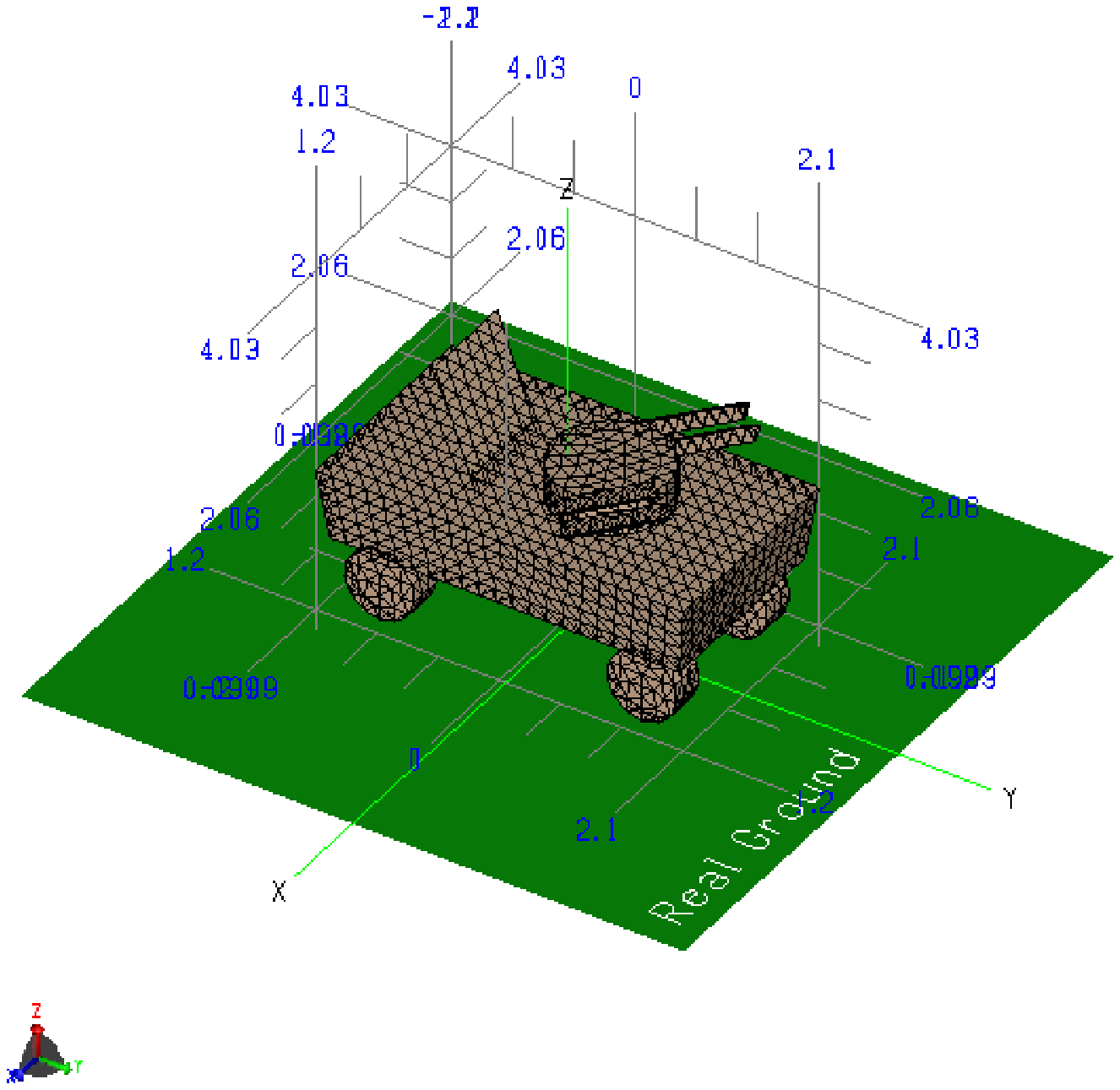} \\ [0.1cm]
\mbox{\bf Missile launcher (MSL)} & \mbox{\bf Stinger (STR)}

\end{array}$
\end{center}
\caption{\label{cad_models}The CAD models of the four ground targets simulated in the present project (dimensions in meters)}
\end{figure*}

Because of the difficulties in obtaining enough field collected data to test the proposed ATR algorithms, an electromagnetic modeling
tool \cite{feko} was used as an alternative to generate a database of bistatic SAR image clips \cite{mishra_09_feko} of
model targets. Figure ~\ref{cad_models} shows the CAD model of the four target used in the simulation. 
The profiles collected were bistatic in nature. 
The transmitter and receiver configuration is shown in figure ~\ref{txrx}. 
For a fixed position of the transmitter, the target is illuminated with the band of frequencies and the surface current is calculated. 
After this, the scattered energy is calculated in a range of receiver directions. 
The bandwidth is centered around 1GHz and the range resolution is 0.3m for monostatic case. 
To keep the range resolution deterioration minimal the bistatic angle, $\beta$, has always been limited to less than $60^o$. 
In the current database we have data for four different types of azimuth angles (varying between $10^o$ to $15^o$) and across the complete $360^o$ of azimuth angle for each elevation angle. 

Some of the limitations of the dataset are as follows. 
The targets are not as complicated as real targets. Hence, the performance figures (in percentage of correct classification) can be taken to show a trend only, not the absolute performance of a practical system.
Secondly, the simulation could only accommodate a flat ground plane. Hence, many of the typical features of radar return data, like ground speckle and shadowing could not be implemented in the data. 
The reader is advised to look at reference \cite{mishra_09_feko} for a detailed report on the simulations steps.

\section{Cognitive  diversity ATR processor}
\begin{figure*}
\centering
	\includegraphics[width=140mm]{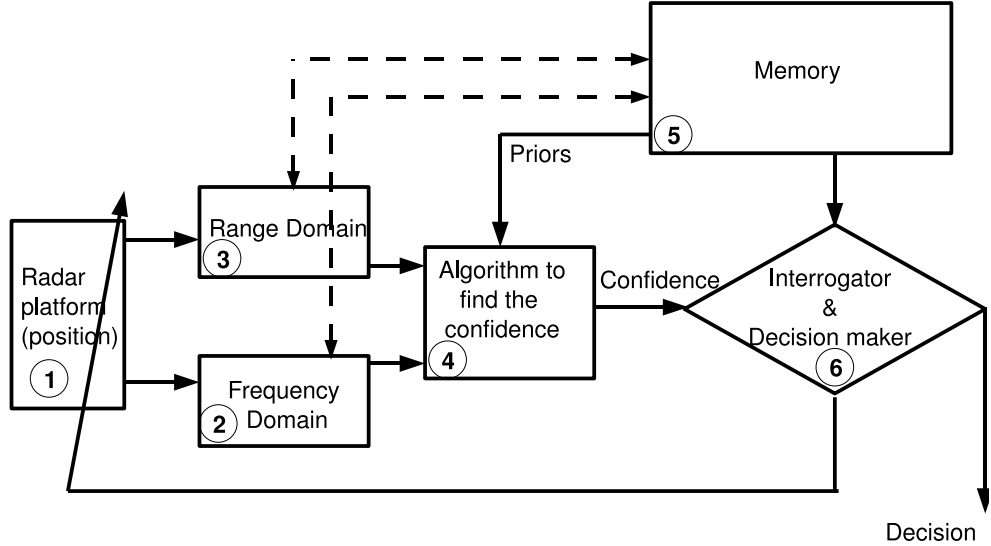}
	\caption{The processing architecture} \label{arch}
\end{figure*}

One of the important observations from the way echo-locating mammals perform in nature is the agility with which they change their position and transmitted waveform depending on their environment and task in hand \cite{haber_83,vespe_10}. 
This has been the major inspiration in the proposed architecture for the ATR machine we will describe in this section. 
A block diagram version of the ATR machine is given in figure ~\ref{arch}. The major blocks (each representing a sub-system of the proposed architecture) are as follows. 
\begin{enumerate}
\item {\it Radar platform:} This block represents the sensor-platform and is also responsible for any preprocessing required for the signal collected by the platform. 
Once the diversity decisions are made, this block has the responsibility to change parameters of operation accordingly. 
It may be noted here that, in the current work, we have dealt with angular diversity only.
\item {\it Range domain processing:} This block handles the processing of the range profile signal as collected from the sensor platform.
\item {\it Frequency domain processing:} This block handles the processing of the data in frequency domain. 
\item {\it Confidence calculation:} This block handles the features and information collected from the above two blocks to make a decision regarding the type of target in the scene along with a confidence with which this decision is made. 
This confidence level representation can be in terms of a continuous variable like probability or in terms of simple discrete levels like \textit{CONFIDENT} or {\it NOT CONFIDENT}.
\item {\it Memory:} Memory description and usage is a crucial part for any automated system endeavouring to become cognitive. 
However, the current memory block only supplied prior test-phase based information to block 4 and block 6.
\item {\it Decision maker:} This block takes the decision regarding whether to go for a fresh collection of data from the scene or not and regarding what diversity to be employed for the fresh data-collection step.
\end{enumerate}

As can be marked, the fifth block is the memory block containing both short  and long term memory. 
From a bio-inspired point of view, the short term memory is coined as {\it echoic memory}, while the long term memory is coined as {\it experimental memory}.

In the current study the system has been validated with three major limitations. 
First of all, we allow only angular diversity. 
So there are two outputs from block 6, viz. decision regarding whether to go for a fresh look of the target or not and by what angle the sensor platform should move from its current position. 
Secondly, the degree of freedom of the sensor platform is limited to only azimuth angle variation. 
The change is azimuth angle prescribed by the decision maker block is represented by $\Delta \theta$. 
A further limitation is that the output of block 4 is discrete in terms of \textit{CONFIDENT} or {\it NOT CONFIDENT}. 
For this decision, a voting method is used while dealing with multiple perspective data in time and frequency domain. 
The decision can be called {\it CONFIDENT} when it is supported by a user-defined percentage of votes. We have taken this percentage to be 50\%.

Algorithm 1 shows the basic steps taken by the proposed ATR architecture.  
For each perspective a predetermined number $N$ of consecutive profiles are used. 
In our study $N$ has been kept limited to 1 while dealing with both the blocks 2 and 3, and $n = 2$ was chosen when the conventional single channel (block 2 or 3) based ATR is studied.

\begin{algorithm}                      
\caption{Algorithm for cognitive angular diversity based ATR framework}          
\label{basic}                           
\begin{algorithmic}[1]
\FOR{$i$ = 1 to $N$}
\STATE {$F_i \Leftarrow$ $k$-space radar signal from block 1}
\STATE {$R_i \Leftarrow$ DFT($F$)}
\ENDFOR
\STATE {Extract features from $F_i$s and $R_i$s}
\STATE {Find the class of the target as predicted by each of the $R_i$s and $F_i$s}
\IF {A single target-class has more than $N+k$ votes}
\STATE {Declare the class with maximum votes as the class of the target}
\STATE {Set the confidence level to {\it CONFIDENT}}
\ELSIF {Number of perspectives used $> K$}
\STATE {Declare the target as unclassified}
\ELSIF {No single target-class has more than $N+k$ votes}
\STATE {Declare the class with maximum votes as the class of the target}
\STATE {Set the confidence level to {\it NOT CONFIDENT}}
\STATE {Go for data collection from a fresh perspective with the platfrom azimuth changed by $\Delta \theta$}
\ENDIF
\end{algorithmic}
\end{algorithm}

With the above base framework, we used a Bayesian decision maker for the {\it cofidence calculation} block. 
%
We term it as {\it Naive Gaussian pdf based Bayesian decision maker}. 
The data is assumed to follow a Gaussian distribution. 
The distribution for different target classes are assumed to have the same covariance matrix and hence the name naive Gaussian pdf based Bayesian decision maker. 
Hence, target variability is provided by the difference in mean only. 
It is also known that radar data vary drastically with aspect angle. 
Hence, instead of assuming the same distribution for a target irrespective of its orientation, we have used 25 pdfs to represent a target as observed from a particular elevation and across the $360^o$ of azimuth angle. 
It may be noted here that this algorithm is the same as the template-matching classifier which has been reported to have very good performance for radar based ATR exercises \cite{novak_99}.

Within this framework, we have tested three types of data processing. 
In the first type, data in time/space domain is handled. This is similar to the conventional ATR processing methods. 
In the second type, both time and frequency domain data is handled simultaneously. 
This will be referred to as time-frequency domain algo 1.
In the third type, decision is tried to reach with the desired confidence using time domain data only. In case no single class get the desired number of votes, then data in frequency domain are processed to bolster the decision making. 
This will be referred to as time-frequency domain algo 2.

\section{Results and Discussions}
The results from the current work will be discussed under two phases. 

\begin{figure}
\centering
	\includegraphics[width=90mm]{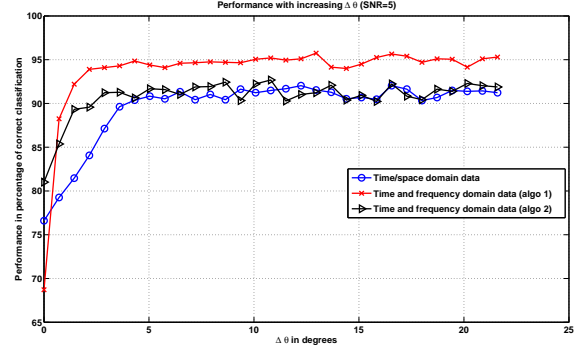}
	\caption{Naive Gaussian pdf based Bayesian decision maker based cognitive diversity ATR} \label{cog1}
\end{figure}

In the first phase, we will check how the algorithms behave with increasing values of $\Delta \theta$. 
Figure ~\ref{cog1} shows the result. 

It can be observed that using both the channels (blocks 2 and 3) gives better performance than using a single channel. 
The conventional algorithm of single perspective based ATR is when $\Delta \theta$ = 0. 
Two of the information not depicted in the graphs are as follows.
For each $\Delta \theta$, the median number of perspectives taken by the algorithms is also noted. 
Irrespective of the value of $\Delta \theta$, the median number of perspectives was found to be 3. 
Secondly, as compared to the proposed variable perspective ATR scheme, if only two perspectives were used with $\Delta \theta = 5^o$, the performance was found to be $85\%$. 
Hence, the cognitive framework performs better than both the single perspective ATR scheme as well as the fixed-number of multi-perspective based scheme.

\begin{figure}
\centering
	\includegraphics[width=90mm]{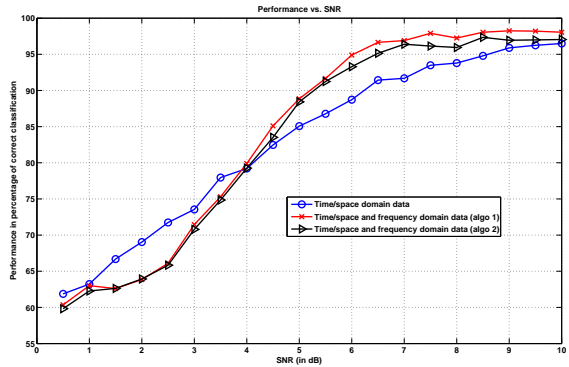}
	\caption{Effect of noise on different cognitive diversity ATR algorithms} \label{snr}
\end{figure}

In the second phase, we analyse how the algorithms behave with different values of SNR figures.
Figure ~\ref{snr} shows the results. For these results, $\Delta \theta$ has been fixed at $3.6^o$. 
It can be observed that the performance of algorithm handling both time and frequency domain data out performs that of handling time/space domain data only, for the cases of high SNR. 

\section{Conclusions}
A cognitive diversity framework based ATR scheme has been presented in this paper. 
A limited version of the scheme was validated using synthetic dataset. 
The performance was found to be better than conventional ATR algorithms of single perspective ATR and multi-perspective ATR usin fixed number of perspectives.
It was also shown that handling the data in both space and frequency domain can enhance ATR performance for higher values of SNR.
 
In the future endeavours, we plan to eliminate the limitations of the scheme and to validate it in a more rigorous way. 
It can be mentioned here that validation of a cognitive ATR scheme against a conventional ATR algorithm may not always be possible. 
This is mainly because the definition of a cognitive system will carry with it a set of conditions and environmental criteria which will not be applicable for conventional ATR schemes.

\section*{Acknowledgement}
Authors thank the Endeavour Research Fellowship Program run by Australian Government's Department of Education, Employment and Workplace Relations, under which the first author was sponsored to work at ANU for a period of six months.
\bibliographystyle{IEEEtran}
\bibliography{mypapers}

\end{document}